\def\year{2022}\relax
\documentclass[letterpaper]{article} 
\usepackage{aaai22}  
\usepackage{times}  
\usepackage{helvet}  
\usepackage{courier}  
\usepackage[hyphens]{url}  
\usepackage{graphicx} 
\usepackage{natbib}  
\usepackage{caption} 
\DeclareCaptionStyle{ruled}{labelfont=normalfont,labelsep=colon,strut=off} 
\usepackage{algorithm}
\usepackage{algorithmic}

\usepackage{booktabs}  

\usepackage{refcount}
\usepackage{bbm}
\usepackage{makecell}
\usepackage{amsmath}
\usepackage{graphicx}
\usepackage{amssymb}
\usepackage{amsthm}
\usepackage{multirow}
\usepackage{xspace}
\usepackage{subcaption}

\usepackage{newfloat}
\usepackage{listings}
\floatstyle{ruled}
\newfloat{listing}{tb}{lst}{}
\floatname{listing}{Listing}

\usepackage{soul}
\usepackage{tikz}
\usetikzlibrary{calc}
\usetikzlibrary{decorations.pathmorphing}

\definecolor{citecolor}{RGB}{34,139,34}
\usepackage[pagebackref=true,breaklinks=true,letterpaper=true,colorlinks,citecolor=citecolor,bookmarks=false]{hyperref}

\usepackage{xspace}

\def\@onedot{\ifx\@let@token.\else.\null\fi\xspace}
\DeclareRobustCommand\onedot{\futurelet\@let@token\@onedot}

\newcommand{\figref}[1]{Fig.~\ref{#1}}

\newcommand{\tabref}[1]{Tab.~\ref{#1}}

\def\ie{i.e.}

\newcommand{\X}{\cal X}
\newcommand{\Y}{\cal Y}

\renewcommand{\[}{\begin{eqnarray}}
\renewcommand{\]}{\end{eqnarray}}
\newcommand{\R}{\mathbb{R}}
\newcommand{\E}{\mathbb{E}}

\newcommand{\N}{{\cal N}}

\newcommand{\ReconstructedPhi}{\phi'(x)}
\newcommand{\CounterPhi}{\psi(x)}

\title{Latent Space Explanation by Intervention}
\author {
    Itai Gat\textsuperscript{\rm 1}\equalcontrib,
    Guy Lorberbom\textsuperscript{\rm 1}\equalcontrib,
    Idan Schwartz\textsuperscript{\rm 1}\textsuperscript{\rm 2},
    Tamir Hazan\textsuperscript{\rm 1}
}
\affiliations {
    \textsuperscript{\rm 1} Technion - Israel Institute of Technology\\
    \textsuperscript{\rm 2} NetApp
}

\usepackage{bibentry}

\begin{document}

\maketitle

\begin{abstract}
    The success of deep neural nets heavily relies on their ability to encode complex relations between their input and their output. While this property serves to fit the training data well, it also obscures the mechanism that drives prediction. This study aims to reveal hidden concepts by employing an intervention mechanism that shifts the predicted class based on discrete variational autoencoders. An explanatory  model then visualizes the encoded information from any hidden layer and its corresponding intervened representation. By the assessment of differences between the original representation and the intervened representation, one can determine the concepts that can alter the class, hence providing interpretability. We demonstrate the effectiveness of our approach on CelebA, where we show various visualizations for bias in the data and suggest different interventions to reveal and change bias.
\end{abstract}

\vspace{-0.05cm}
\section{Introduction}

Machine learning is ubiquitous these days, and its impact on everyday life is substantial. Supervised learning algorithms are instrumental to autonomous driving~\citep{lavin2016fast, bojarski2016end, luss2019generating}, they serve people with disabilities~\citep{tadmor2016learning}, they are applied to improve hearing aids~\citep{Hurtley645, schroter2020clcnet, fedorov2020tinylstms}, and they are being extensively used in medical diagnosis~\cite{deo2015machine, qayyum2020secure, seo2020machine, richens2020improving}. These advancements are achieved with complex models, and their decisions are usually not well-understood by their operators. Consequently, model interpretability is becoming an important challenge for contemporary deep nets.

To this end, a magnitude of interpretability methods has been proposed. Current model interpretability methods are geared towards local feature relevance. In other words, given a sample, they quantify how much each feature contributes to the prediction. For example, gradient-based approaches are widely used to interpret the model predictions since they bring forth an insight into the internal mechanism of the model~\citep{saliency, gradcam, integrated, deeplift, guidedbackprop}. These gradient-based methods produce different explanation maps using the gradient of a class-related output with respect to its input data. 
While relevance scores in the input sample are human-interpretable, they are sparsely scattered across many pixels and do not accurately indicate concepts. Recently, counterfactual examples were used to measure the causal effect of known data concepts~\citep{feder, atzmon, causalm, acl_rosen}. However, these methods require costly predefined concepts annotation, which is rarely apparent when learning from data.

\begin{figure}
\includegraphics[width=0.46\textwidth]{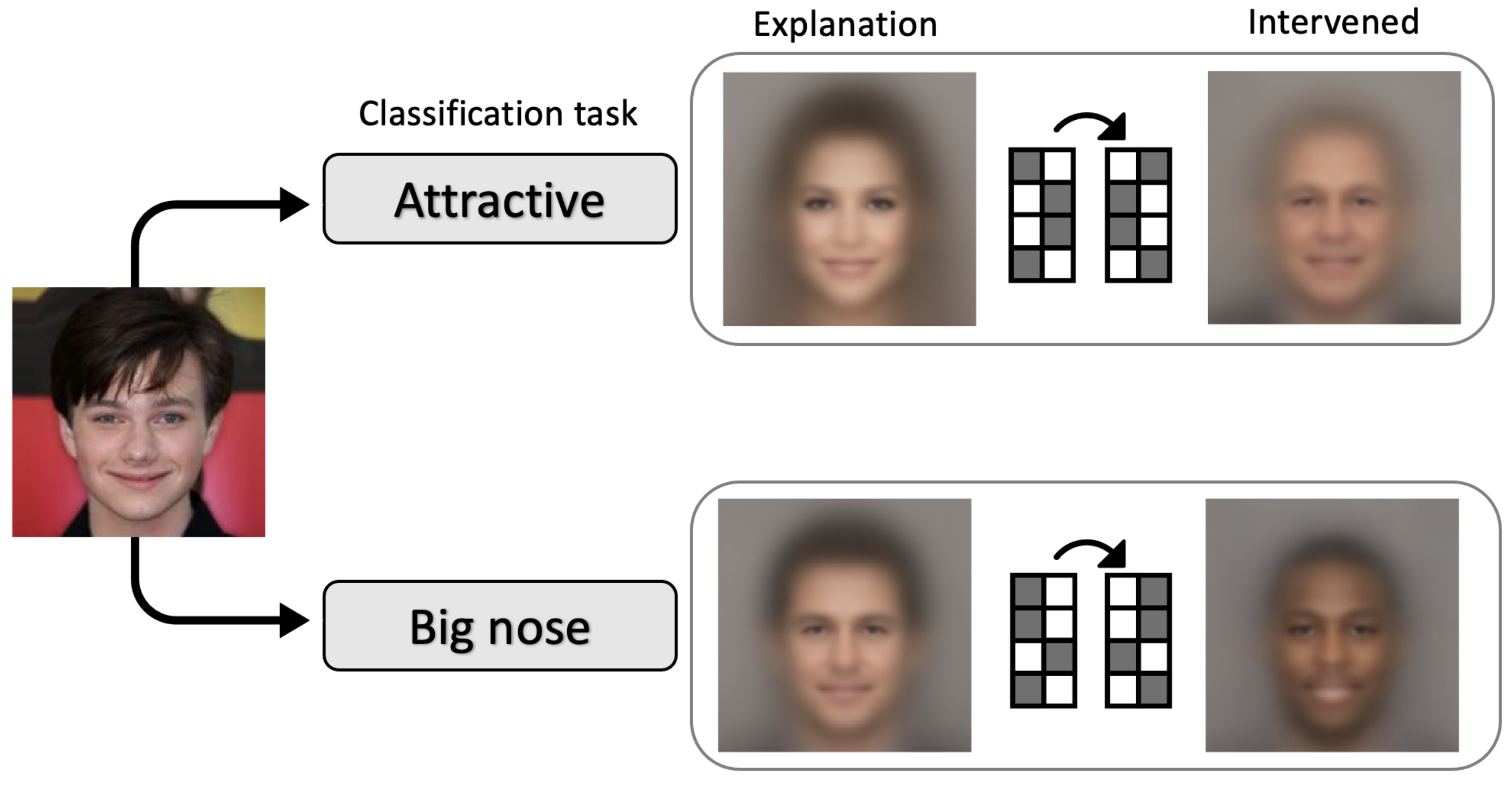}
\centering
\caption{Our method explains a layer of a given discriminative learner. We learn a discrete variational autoencoder of the last layer representation over the data. Then, we produce a visual explanation of it. Using the discrete latent space, we can intervene in the conceptual world of the data, which allows us to ‘see’ the world from the viewpoint of the classifier. This figure illustrates the explanation for the last layer in two classification tasks on the CelebA dataset: attractive and big nose. We find out that the classifier attributes attractiveness to young women and older man features to the opposite label. This could imply a bias in the annotation process of CelebA and/or reflects an inherent bias in the data.}
\label{fig:teaser}
\vspace{-0.5cm}
\end{figure}
In this work, we focus on finding the high-level concepts that explain the classifier's decision. The bias within these concepts is often the result of overlooked priors that are woven into the dataset. In order to reveal bias in the data, we propose a Variational Auto-Encoder (VAE) mechanism that discretizes the input representation. This discrete latent space is able to capture the concepts that are embedded in the representation and allows us to avoid costly annotation of predefined concepts that impede the utilization of modern causal effect techniques. Using an intervention mechanism that flips the boolean latent space representation bits, we can infer the concepts that were responsible for the prediction and detect bias in the data. 

We take advantage of recent developments in the field of generative learning in order to reconstruct human-friendly explanation images for a discrete latent space concept. Based on this reconstruction, the guiding concepts can be identified. For instance, in \figref{fig:teaser}, concepts are illustrated abstractly. Following the intervention, examining the shifted concepts, we identify the relevant factors. For instance, age and gender appear to influence attractiveness, while skin color seems to influence a big nose's attractiveness.  In addition, we optimize shared information between the visual explanation and classifier in order to ensure that the abstract concepts rely on the same information as the discriminative classifier. This regularization strategy increases the corresponding information and ensures that the realized concepts correspond to the classifier's guiding concepts.

The remainder of the work is organized as follows: In the method section, we present our approach for learning discrete concepts of a hidden layer representation. It allows us to intervene and change the data concepts. We then propose to generate visualizations of the discovered concepts by an explanatory approach. We conclude the method section by providing a theoretic framework for our method via the functional Fisher information and use it as a regularization term. In the results section, we study our method qualitatively and quantitatively on different tasks on CelebA dataset. We first illustrate our intervention mechanism for concept discovery and manipulation. Then, we use our method to detect biases in the discriminative model. We then verify our approach with quantitative and subjective studies.

In summary, our contributions in this work are:
\begin{itemize}
    \item Our method identifies discrete global concepts in a trained layer. Furthermore, it enables to intervene on those global concepts.
    \item We propose a novel explanatory  approach for explaining a given layer of a trained classifier in a human interpretable manner.
    \item We integrate discriminative and explanatory  functional information into a single information-theoretic framework. We then introduce a regularization term that encourages high shared information between discriminative and explanatory  learners.
\end{itemize}
We study the effectiveness of our approach in generating explanations, detecting bias, and controlling concepts for different tasks on high-dimensional images.

\section{Related work}

\vspace{-0.05cm}
\subsubsection{Explainability in computer vision}

Explainability takes many forms and has not been formally defined in the literature.  In general, the methods can be divided into two branches: global explanations (such as a decision tree's structure) and local explanations (such as important features of samples). Among deep learning explanations, local explanations predominate. The reason for this is that the backpropagated gradients naturally provide a sample with a heatmap that shows a sample's features' relevance. Early attempts created saliency maps based on the gradients~\cite{saliency, dabkowski2017real, mahendran2016visualizing}. Following that, a variety of a mixture of gradient and input methods emerged~\cite{gradcam, deeplift, smoothgrad, srinivas2019full}. Attribution propagation approach calculate relevance scores based on a set of axioms~\cite{bach2015pixel, montavon2017explaining, nam2020relative, deeplift, gu2018understanding, chefer2021transformer}.  Attention models also provide explainability~\cite{fga, schwartz2017high, simple-avsd,  braude2021towards, ali2022video}. Another approach, deconvolution networks, are capable of providing insights into intermediate layers by calculating the transposed convolutional network~\cite{zeiler2014visualizing}. However, further research suggests that deconvolution is similar to gradient backpropagation~\cite{guidedbackprop}. These methods are similar to ours in that they use reconstruction to find relevance. By contrast, our approach reconstructs a concept image to reveal a high-level concept rather than pixel relevance. \citet{li2018patternnet} study the ability of retrained networks to predict relevance maps. Among other popular methods, LIME approximates a model's prediction with a local linear function~\cite{lime}.  \citet{shap}  propose SHAP, a unified game-theoretic framework for attribution methods based on Shapley values~\cite{shapley}. The computational complexity of such methods is considerable, and their efficiency is often not as high as that of other methods. ~\citet{frosst2017distilling,wan2020nbdt} examine networks that are explained by the architecture design. Notably, all of the above methods investigate the pixel relevance of individual data points.  However, our work seeks to reveal the underlying high-level concepts, and so is also infused with global explanations. A global explanation is favored in assessing the robustness of a model. E.g., it can uncover a global biased concept (e.g., smiling classification is affected by gender)~\cite{doshi2017towards}. Our reconstruction is applied to the deep layers, which typically contain information related mainly to labels, suggesting a global nature~\cite{information2}. Hence, it is concise in its representation of the task and does not include all the concepts that pertain to the sample.

\begin{figure}
     \centering
     \includegraphics[width=0.45\textwidth]{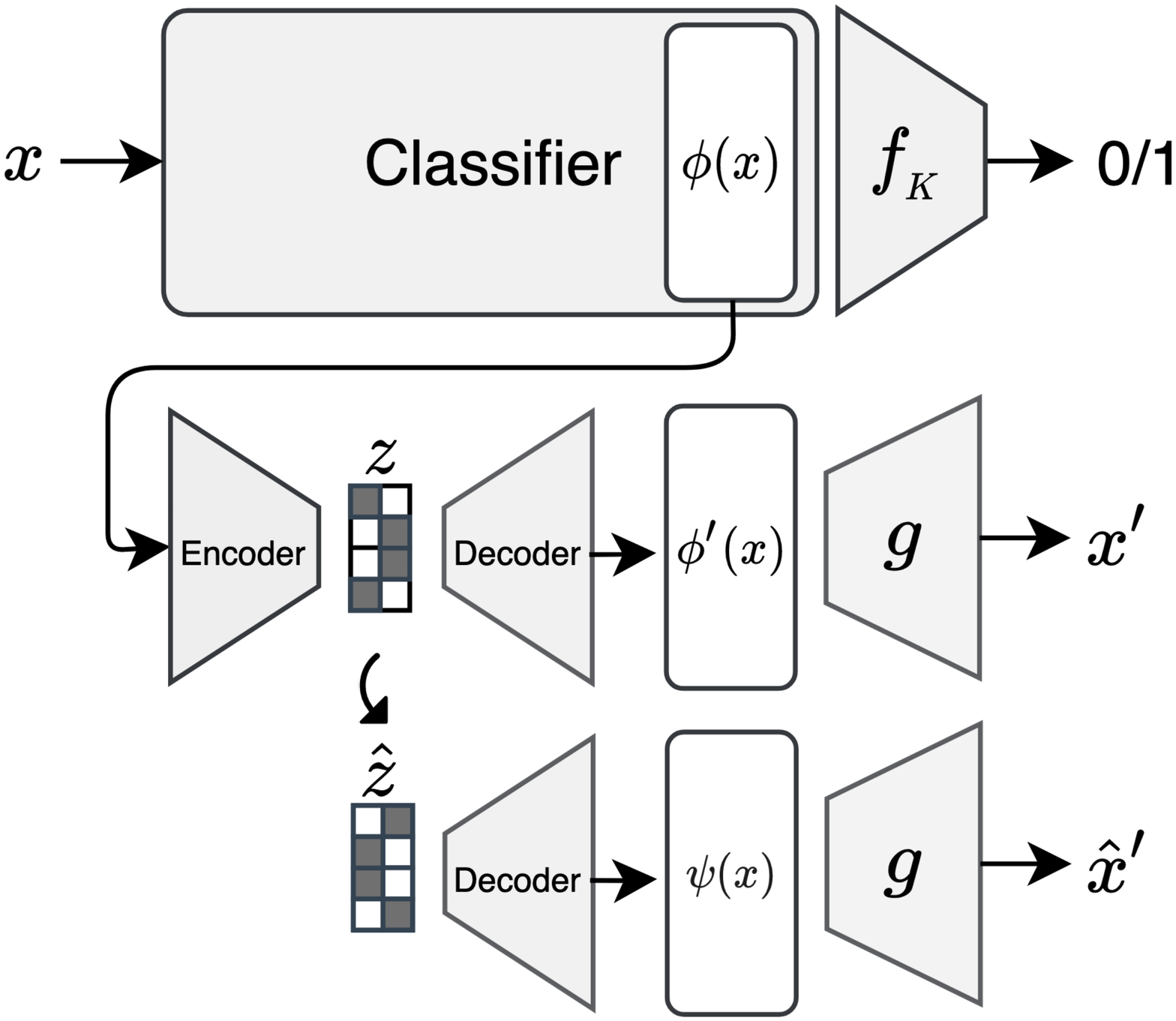}
     \caption{An overview of our approach: Our method provides visual explanations of a given layer of a trained classifier. By using a discrete variational autoencoder, we learn a boolean representation $z$ of $\phi(x)$ that encodes the concepts that drive the classifier prediction. Then, to visualize these concepts a generator is trained to reconstruct the original image given $\ReconstructedPhi$. By intervening in $z$ and getting a modified version of $\ReconstructedPhi$ $\CounterPhi$, one can evaluate the qualitative differences using $g(\CounterPhi)$ and the quantitative differences using $f_{_K}(\CounterPhi)$. This intervention mechanism provides the ability to interpret the classifier prediction.
     }
    \label{fig:three graphs}
    \vspace{-0.3cm}
\end{figure}

\vspace{-0.05cm}
\subsubsection{Intervention for causal discovery} 

In recent years, a mainline of research has focused on quantifying the causal effect of data concepts on prediction. The simplest form of intervention are perturbations, i.e., removing and adding pixels to generate counterfactual local explanations for images~\cite{fong2019understanding, fong2017interpretable, feder}. These approaches can also be used to explain models without looking at their mechanics, i.e., a black-box approach. Despite their appealing black-box nature, there are many pixels in a picture, and each pixel can have any value. Therefore, manipulating the combination of pixels can be computationally complex. Our work is different in that it seeks to find high-level concepts by transforming latent space into discrete boolean space using a Variational Auto-Encoder (VAE). As a result, the space of possible interventions is significantly smaller. To be more precise, we study a single flip that shifts the predicted class as the binary values change. Also relevant to our work are causality frameworks, such as TCAV~\cite{pmlr-v80-kim18d} and CaCE~\cite{feder}. These methods intervene in labeled concepts to measure the causal effects. However, they require data with labeled concepts. This kind of annotation process must acquire the expertise to identify the underlying structure of the problem, and even then, there is still the human bias that contaminates the process. Using our method, we uncover the task's decisive concepts without labeling them. A concepts image is reconstructed using deep layer representations. Note that deep layers typically contain label-related information~\cite{information2}. The consequent advantage is that it is easy to pick up general concepts related to labels. Once we intervene in the concepts images, decisive concepts emerge (see Fig.~\ref{fig:teaser}).

\vspace{-0.05cm}
\subsubsection{Bias in datasets} 

Also relevant to our work is bias in datasets. The ImageNet challenge~\cite{imgnet} and the development of AlexNet~\cite{alexnet} sparked the deep learning revolution. However, as datasets grow, biases emerge, which may go undetected for a long time. In ImageNet, for instance, the background can indicate information about the class of objects~\cite{inception}. Multi-modal models only use one modality due to dataset priors~\cite{gat2021perceptual}. In addition to this, a number of datasets are being annotated in uncontrolled environments, as crowdsourcing services such as Amazon Mechanical Turk become more and more popular~\cite{geva2019we}. For instance, well-known benchmark `Faces in the wild'~\cite{faces_in_the_wild} contains 70\% male and 80\% white-skinned faces.  In our study, we examine a manipulated dataset where we altered the class gender majority to demonstrate that our method identifies hidden factors that affect predictions (see Fig.~\ref{fig:dataset_bias}).

\vspace{-0.05cm}
\section{Method}\label{sec:method}

Learning a discriminative model amounts to fit function $f: \X \rightarrow \Y$ from the space of data $x \in \X$ to the space of labels $y \in \Y$ using parameters. The parameters of the function $f(x)$ are learned by fitting to a training data $\{(x_1,y_1),..., (x_m,y_m)\}$ using a loss function $\ell(f(x_i),y_i)$. 

Whenever $f(x)$ is a linear function, e.g., in the case of logistic regression and support vector machine, one can explain the importance of features $x$ by the magnitude of the learned coefficient parameters. Unfortunately, this is not the case when $f(x)$ is learned by a deep net. 

Deep nets are able to learn complex relations from $x$ to $y$ to their recursive structure. For instance, consider a  fully connected deep net, where the input vector of the $k$-th layer is a function of the parameters of all previous layers, i.e., $\phi_k(x)$. The entries of $\phi_k(x)$ are computed from the response of its preceding layer, i.e., by the linear relation $W_{k-1} \phi_{k-1}(x)$, followed by a non-linear transfer function $\phi_k(x) = \sigma(W_{k-1} \phi_{k-1}(x))$. While this recursive structure allows to easily encode complex functions $f(x)$, it also renders the model decisions hard to explain. Specifically, we are interested in the last hidden representation since it primarily encapsulates label-related information~\cite{information_bottleneck}. For notation clarity, we denote it as $\phi(x) = f_{0:K-1}(x)$, where $K$ is the last hidden representation. We also denote $f_K$ as the remaining last layer of the classifier. 
In the following, we propose a two-stage explanation of recursive representations of a deep network. In the first stage, concepts are detected using a clustering over $\phi(x)$ (see Sec. \textit{Discrete concepts in a hidden layer of a discriminative model}). In the second stage, we produce visual explanations using an intervention mechanism (see Sec. \textit{Visual explanation of a latent space}). The generated images can be compared to the original data point $x$ under different interventions. The entire flow of our approach is illustrated in 
\figref{fig:three graphs}. The final section introduces a regularization term, which encourages the generator to use the same information as the classifier to illustrate the concepts used by the classifier accurately (see Sec. \textit{Maximizing explanatory  and discriminative shared information}). In the experimental validations, we demonstrate how such interventions can reveal bias in the data. In the following, we begin by introducing our intervention mechanism.

\vspace{-0.05cm}
\subsection{Discrete concepts in a hidden layer of a discriminative model}
\label{concepts_detection}

Our goal is to reveal and visualize the underlying concepts that drive the model's prediction. Broadly, a classifier encodes information in a hidden layer in order to discriminate between the different classes of the label space. Thus its hidden layer representations correlate with the relevant target classes. For example, hidden layers may encode body parts when learning to recognize a person. 

Unfortunately, the continuous nature of a hidden layer of the discriminative model does not allow us to reveal these concepts easily. For this purpose, we employ a discrete variational autoencoder (DVAE) with boolean latent variables $z \in \{0,1\}^n$~\cite{rolfe2016discrete}. In our case, the DVAE learns discrete concepts $z$ of its input vector $\phi(x)$. This input vector is a hidden layer of the discriminative model whose discrete concepts we want to reveal.  The DVAE learns these concepts by maximizing the likelihood $p(\phi(x))$ using the negative evidence lower bound (ELBO):  $-\log p(\phi(x)) \le$ 
\begin{small}
\[
- \E_{z \sim q(\cdot | \phi(x))} \Big[\log p(\phi(x) | z) \Big] - KL \Big(q(z | \phi(x)) || p(z) \Big),
\]
\end{small}
where $q(z | \phi(x))$ is a probability distribution of the $n$ dimensional boolean variable $z \in \{0,1\}^n$. We use the Gumbel-softmax reparametrization  for estimating the encoder's gradients through the non-differentiable boolean layer~\cite{maddison2016concrete, jang2016categorical}.

We denote by $\ReconstructedPhi$ the reconstructed embedding vector (see \figref{fig:three graphs}). The boolean latent space $z = (z_1,...,z_n)$ enables to reveal discrete concepts of $\phi(x)$. The generative nature of the DVAE allows us to turn turn-off or turn-on some concepts from $z = (z_1,...,z_n)$ using an intervention $\hat z = (\hat z_1,...,\hat z_n)$. This interventions mechanism allows us to modify the generated representation of $\ReconstructedPhi$ to the counterfactual representation $\CounterPhi$.
Although the DVAE allows us to identify the concepts that the discriminative classifier relies on, it is unclear what these concepts are. To visualize these concepts, we produce human interpretable explanations by generating visual explanations derived from a given latent representation $\ReconstructedPhi$.

\begin{figure*}[t!]
\includegraphics[width=0.79\textwidth]{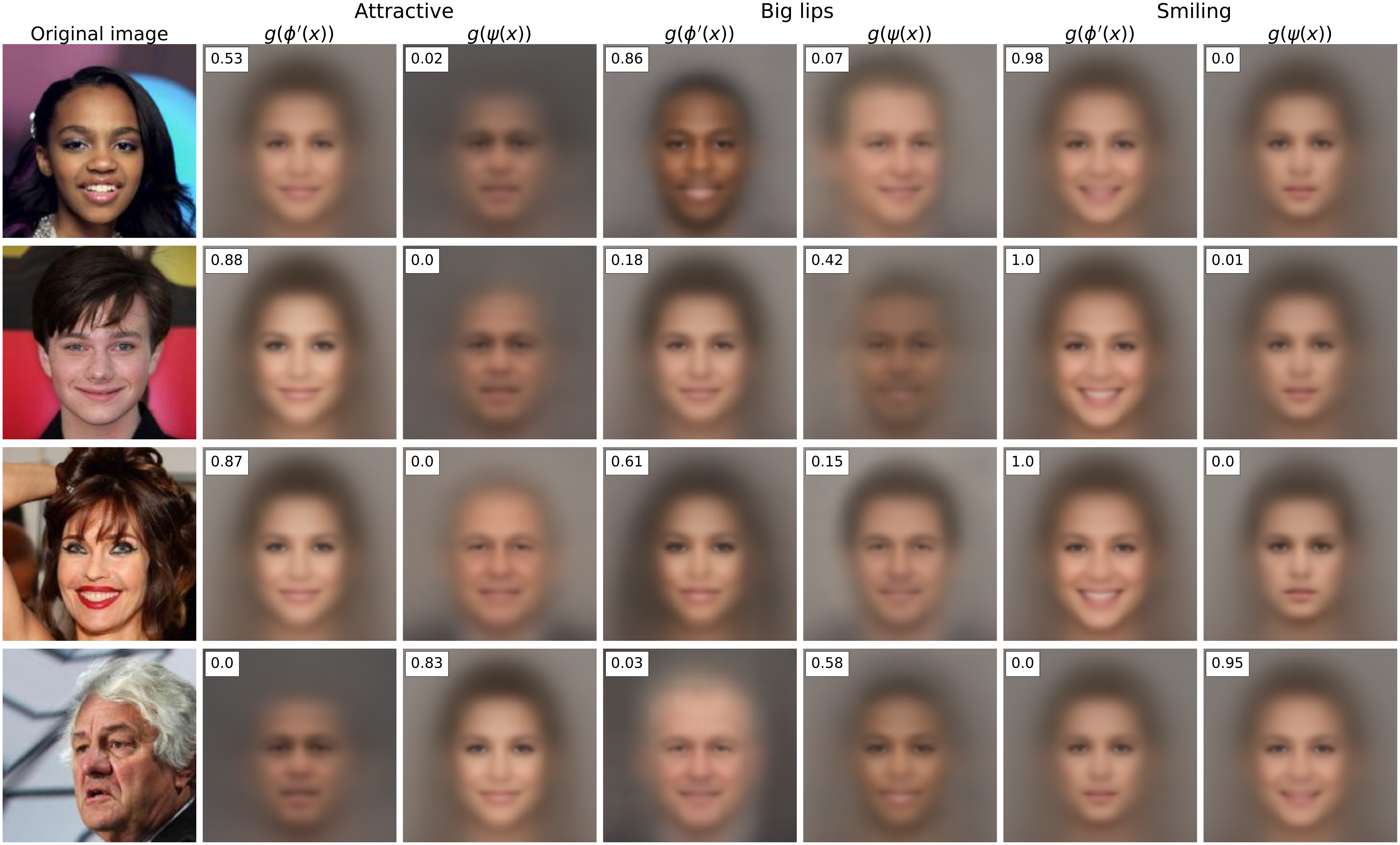}
\centering
\vspace{-0.1cm}
\caption{Explanation by intervention: we present results from our method on three different binary tasks (attractive, big lips, and smiling). For each task, we present our method's explanation on the representation of the original input ($g(\ReconstructedPhi)$) and the flipped discrete representation ($g(\CounterPhi$). In addition, we present the probability that the classifier predicted for the two opposite representations. We note that our methods capture biases in the dataset effectively. Moreover, we forward the flipped representations to the classifier and find that the classifier predicts the opposite class for all input samples.}
\label{fig:counterfactual}
\vspace{-0.4cm}
\end{figure*}

\vspace{-0.05cm}
\subsection{Visual explanation of a latent space}
\label{sec:vis_explain}
\vspace{-0.05cm}

A trained DVAE allows us to learn latent concepts $z \in \{0,1\}^n$ that govern the hidden layer representation $\phi(x)$. Our goal is to intervene and change latent concepts, from $z$ to $\hat z$, and consequently to change the DVAE decoding from $\ReconstructedPhi$ to $\CounterPhi$ in order to detect bias of the discriminative learner. To visualize this bias, we suggest to learn the explanatory function $g: \R^d \rightarrow \R^{d_c\times d_h\times d_w}$, where $d$ is the $\phi(x)$'s embedding dimension and $d_c\times d_h\times d_w$ is the generated image's embedding dimension. Thus, $g$ produces a visualization of the discrete concepts in the latent space using the $x'$ from the hidden representation $\ReconstructedPhi$. Intuitively, the role of $g$ is to output a visual explanation. We learn $g$ to minimize a reconstruction loss of the original image,  
\[
    \ell(g(\ReconstructedPhi), x).
\]
Visualizing the counterfactual representation $g(\CounterPhi)$ allows us to `see' through the eyes of the layer we investigate (see Fig.~\ref{fig:teaser}). Note, training the DVAE and the explanatory network together might interfere with the DVAE's discrete structure learning processes with features related to the image. This is undesirable since the DVAE's role is to learn the representation $\phi(x)$ and not input-related features. Thus, we first train the DVAE and then train $g$ while the DVAE is fixed.

By explaining a latent representation of a classifier $f$ through a learning process, it is not guaranteed that $g$ relies on the same information as the classifier. To encourage this, we propose to measure their shared functional information and maximize it during the learning process of $g$. In the next section, we connect the notion of functional information to our settings and then present a regularization term that maximizes the shared information between the explanatory  network and the discriminative learner. 

\vspace{-0.05cm}
\subsection{Maximizing explanatory  and discriminative shared information}\label{sec:information}
\vspace{-0.05cm}

Our next goal is to ensure that our generated concepts follow the same concepts the discriminator employ. We achieve this by maximizing the amount of information that the explanatory  learner (\ie, $g$) extracts from the latent representation with respect to the discriminative learner's (\ie, $f_K$) information.  

To measure information we follow ~\citet{gat2020removing} and consider the functional information that is encapsulated in a non-negative function $h(z)$ over a Gaussian space $\nu = \N (\mu, 1)$,
\[
{\cal I}_\nu(h) &\triangleq&  \E_{z \sim \nu} \Big[ \frac{h'(z)^2}{h(z)} \Big]. \label{eq:inf}
\]
The functional Fisher information is computationally appealing as it can be easily evaluated using sampling over a Gaussian space.

Based on this measure, we compute the information of our explanatory  function $g(z)$ with respect to the discriminative layer $\phi(x)$ over the explanatory  distribution $\N (\phi(x), I)$.
Since $\N (\phi(x), I)$ is a multivariate Gaussian, we estimate the information with respect to each of its coordinates independently. We denote by $\nu_i$ the univariate Gaussian measure $\nu_i = \N (\phi^i(x), 1)$, where $\phi^i$ is $\phi$'s $i$-th element. We thus define the information in each neuron $i$ in any given layer by 
\[
{\cal I}^i_\nu(g) = \E_{z \sim \nu_i} \Big[ \frac{g'(z_i^\phi)^2}{g(z_i^\phi)} \Big],
\]
where $z_i^\phi = (\phi^1(x),...,\phi^{i-1}(x), z, \phi^{i+1}(x),...,\phi^{d})$. Then, a layer information defined as
\[
{\cal I}_\nu(g) \triangleq ({\cal I}^1_\nu(g),...,{\cal I}^d_\nu(g)). \label{eq:I}
\] 
Next, we integrate the functional information into the explanatory  learning process. We repeat the same procedure for the discriminative function $f_K(\cdot)$ over the same intermediate layer $\phi(x)$. We augment the discriminative layer $\phi(x)$ with a Gaussian space and define ${\cal I}_\nu(f_K)$ as in Eq. (\ref{eq:I}). 

Finally, we encourage the generator to rely on the same information as the discriminative learner. To this end, we add a regularization term to the generator training process. The regularization incorporates a differentiable similarity between the information of the generator. ${\cal I}_\nu (g)$ and ${\cal I}_\nu (f_K)$ where the input of $g$ is $\ReconstructedPhi$ and the input of $f_K$ is $\phi(x)$. 
We employ dot product for similarity and take the inverse in order to account for both reconstruction loss minimization and similarity maximization~\cite{similarity_gradients, task2vec}. This leads to our learning objective, 
\[
    \ell(g(\ReconstructedPhi, x) + \lambda({\cal I}_\nu (g)\cdot {\cal I}_\nu (f_K)^\top)^{-1},
\]
where $\lambda$ is a hyperparameter that calibrates the training loss and the inverse similarity.

\vspace{-0.05cm}
\section{Results}\label{sec:expreiments}

We propose an explanatory approach for latent space explanations using an intervention mechanism. This section studies our explainability method over different classification tasks. We show qualitatively that our intervention mechanism reveals different concepts for different tasks (see \textit{Counterfactual representation}). We then show that our approach reveals deliberately integrated gender bias (see \textit{Dataset statistics control}). To further verify that the classifier is utilizing the detected concept, we verify the prediction of counterfactual samples. To this end, we employ a state-of-the-art generative model and examine the prediction of manipulated images according to the revealed bias (see \textit{Counterfactual verification}).  In addition, we provide two quantitative measures: 1) We assess the quality of our intervention by measuring the number of times the intervention affected the classifier predictions (see \tabref{tab:counterfactual}).  2) We investigate the performance of our proposed regularization term by studying the generator, and the classifier shared information (see \tabref{tab:cosine}).

\vspace{-0.05cm}
\subsubsection{Experimental setup} 
\vspace{-0.05cm}

We performed our experiments on the CelebA dataset~\cite{liu2015faceattributes}, which is annotated with 40 binary attributes. We used its binary attributes as labels for different classification tasks (e.g., attractive, big lips, smiling). For each task, we trained a different discriminative learner (i.e., $f$). Note that the data splits are not necessarily balanced. For example, the big lips attribute is a minority class (i.e., 48,785 vs. 113,985). The discriminative learner is composed of six convolutional layers followed by two fully connected layers. $\phi(x)$ is the output obtained after the first fully connected layer. For the DVAE, the encoder is composed of four linear layers. The decoder is composed of the transposed layers of the encoder. For the reconstruction loss of the DVAE, we used mean squared error loss, and for the reconstruction of the explanatory network, we used binary cross-entropy. In all networks, we used the ReLU activation function and Dropout. We used the same architectures for all the tasks. 

\begin{figure*}[t!]
    \includegraphics[width=0.75\textwidth]{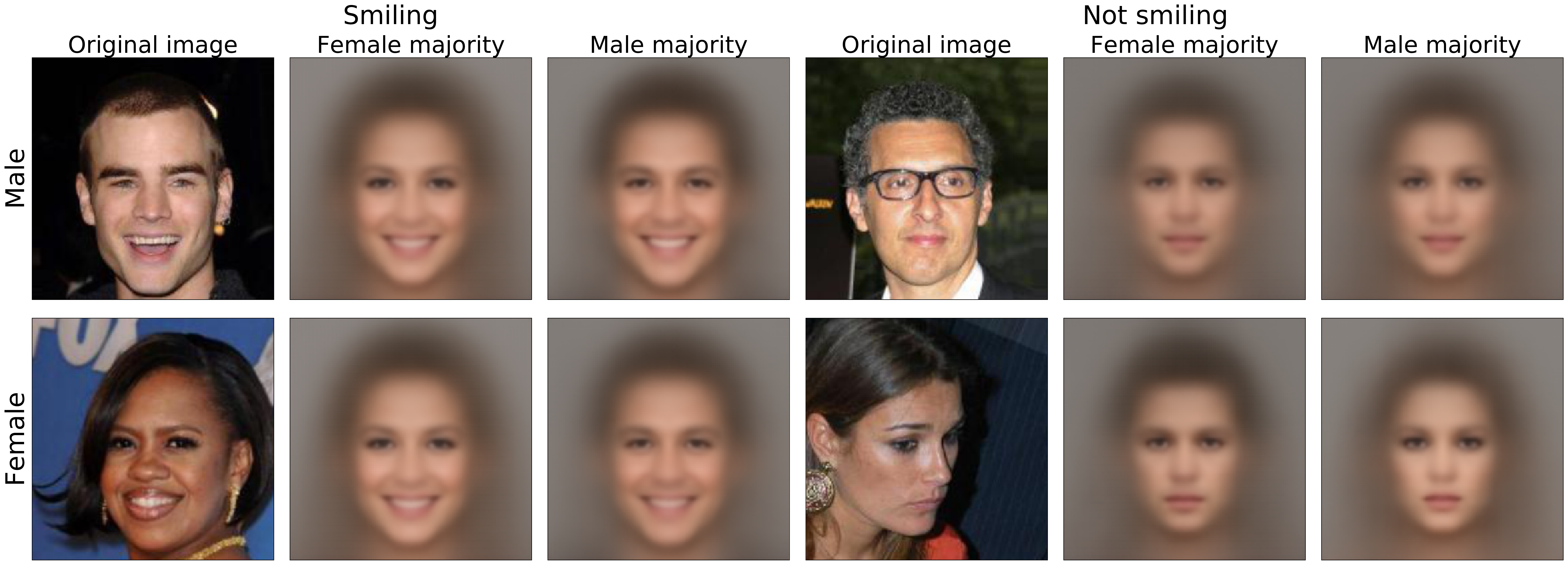}
    \centering
    \caption{An illustration of our method applied to two classifiers trained on different splits. Each split with a different gender majority. A majority of the females are smiling in the first split, while most males are smiling in the second. The results show how male and female concepts are encoded: on the left, when the original images are smiling, the explanations of the female-majority classifier appear feminine, while the male majority appear masculine. On the right, when the original images are not smiling, the explanations are the opposite, i.e., the female-majority classifier appears masculine, while the male majority appears feminine. These findings demonstrate that existing biases in the data, and consequently in the model, emerge in the explanations obtained in our method.
    }
\label{fig:dataset_bias}
\vspace{-0.3cm}
\end{figure*}

\begin{table}[t]
\centering
\resizebox{\linewidth}{!}{%
    \begin{tabular}{lcccc}
    \toprule
        Task & \% of change & Acc. $\phi(x)$ & Acc. $\ReconstructedPhi$ & Agreement \\
        \hline
        Smiling    & 98.86 & 92.10 & 91.80 & 98.08\\
        Attractive & 97.08 & 79.08 & 78.66 & 95.98\\
        Big lips   & 92.13 & 85.20 & 80.33 & 89.70\\
        \bottomrule
    \end{tabular}
}
\vspace{-0.1cm}
\caption{To evaluate the quality of the DVAE for our purposes, we suggest two metrics. The first considers the ability of our method to perform a boolean flip in the prediction. The measure quantifies the percentage of cases in which the prediction changes when the discrete representation is flipped. We observe that, for all tasks, discrete flip changes the prediction to the opposite label. The second evaluates the reconstruction quality. For this, we compute the accuracy for $\phi(x)$ and $\ReconstructedPhi$. We also report the percentage of times the predictions agree on the same label.}
\label{tab:counterfactual}
\vspace{-0.5cm}
\end{table}

\vspace{-0.1cm}
\subsection{Counterfactual representation}\label{sec:exp_explaination_by_generation}
\vspace{-0.05cm}

In the following, we study the ability of our mechanism to identify discrete concepts used by the classifier. In Fig.~\ref{fig:counterfactual}, we present our method on various binary classification tasks such as attractiveness, big lips, and smiling (more studies presented in the supplementary material). We show that the concepts images (i.e., $g(\ReconstructedPhi)$) depict similar abstract representations regardless of the data input. The reason for this is that deep layers only contain information specific to downstream work. As a result of the narrowing of information, the generator must compensate for the lost information to reconstruct the original image. Thus, the shown concepts are mostly associated with the downstream task label. For instance, the first three images of the second column all portray a young woman with feminine features, despite the fact that the three individuals are of different ages, skin colors, and sex. 
Next, we will achieve counterfactual representation through intervention. Note, this intervention is possible due to the use of binary representation. Our study illustrates interesting concepts that affect the prediction: 1) In the second and third columns, we examine attractiveness and discover that counterfactual samples of attractive persons display an older figure. 2) In the fourth and fifth columns, we examine the big lips classification and conclude that intervention produces different skin colors. 3) In the sixth and seventh columns, we examine the smiling classification and find that the gender characteristics flip once the intervention is conducted. Note, each user has the option of determining whether or not these concepts constitute biases. Our goal is to expose which concepts the classifier might pick, not to determine whether or not they are biased. For instance, whether age can be used to classify attractiveness is at the discretion of the user. We next propose metrics to ensure our DVAE works properly.

In Tab.~\ref{tab:counterfactual},  we measured the accuracy when we inject the reconstructed representation $\ReconstructedPhi$ to $f_K$ (i.e., $f_K(\ReconstructedPhi)$).
We tested similarities between the reconstructed representation ($f(\ReconstructedPhi)$) and the original representation ($\phi(x)$). We found that the reconstructed representation keep most of the original accuracy performance. For instance, for the smiling task, the accuracy of the reconstruction representation is 91.80\%, while the original accuracy is 92.10\%. We further suggest to quantify the quality of the intervened representation by the percentage of input data points that change their prediction. We show that our method flips the prediction when we employ the flipped discrete representation (i.e., $\CounterPhi$). We found 98.86\%, 97.08\%, 92.13\% percent of the samples flipped prediction for the smiling, attractive, and big lips classification respectively.

\begin{figure}[t!]
    \includegraphics[width=0.45\textwidth]{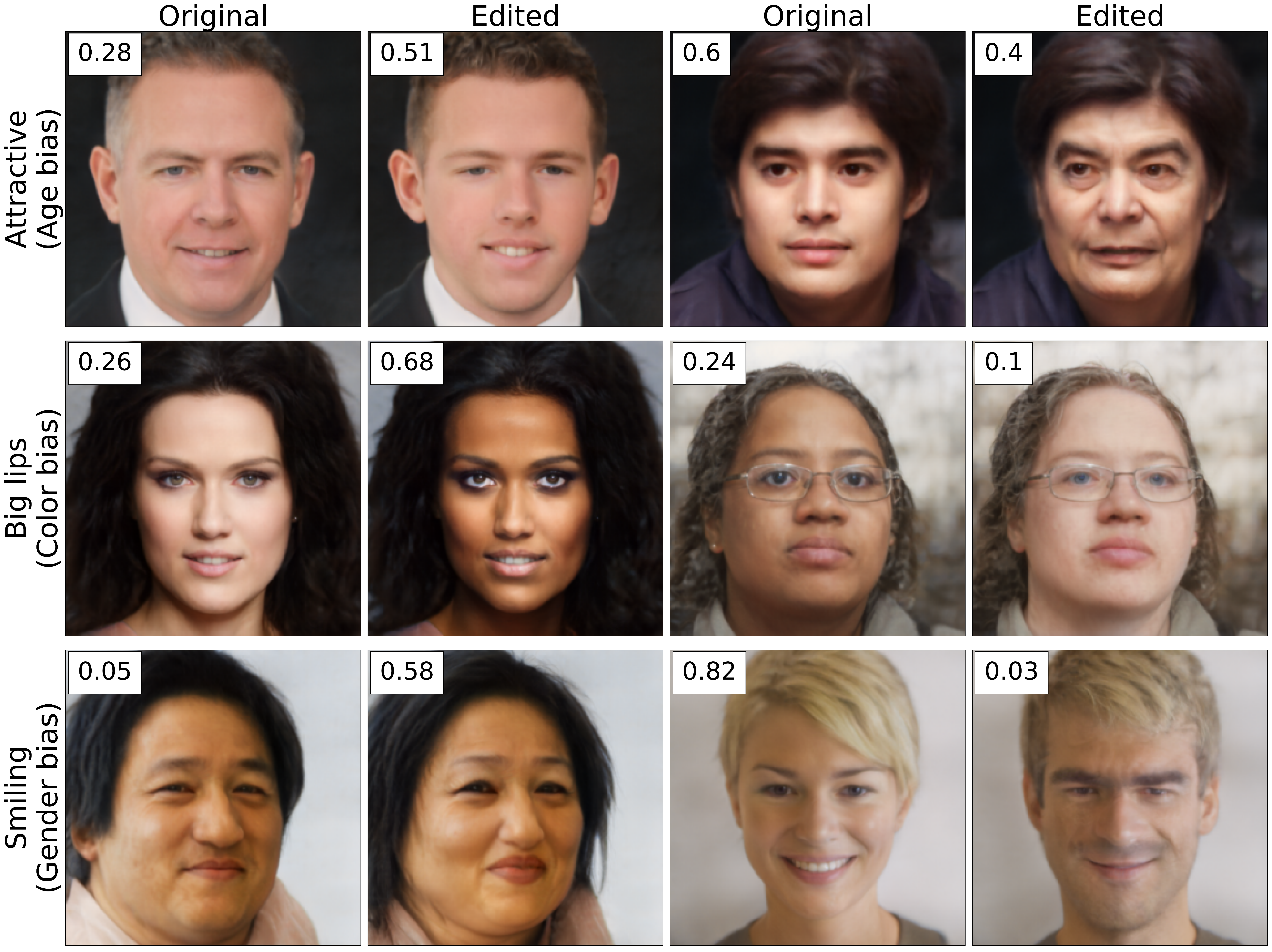}
    \centering
    \vspace{-0.2cm}
    \caption{We edited input images to create their counterfactual biased versions in order to test whether the bias concepts detected by our method affect the classifier predictions. Every row represents a different classification task. The first and third columns show the original images with the absence and presence of the label attribute, respectively. In the second and fourth columns, we show the edited image. In addition, we show the prediction score for both the original and edited images. For instance, in the second row, the skin color affects the predicted scores for the big lips task (the prediction score is 0.26 vs. 0.68).}
    \label{fig:clip}
    \vspace{-0.4cm}
\end{figure}

\begin{table}[t]
\centering
\resizebox{\linewidth}{!}{%
    \begin{tabular}{lcc}
        \toprule
        Task   & Without regularization &  With regularization \\
        \hline
        Smiling    & 0.43 & 0.84 \\
        Attractive & 0.37 & 0.80 \\
        Big lips   & 0.51 & 0.85 \\
        \bottomrule
    \end{tabular}
}
\vspace{-0.2cm}
\caption{We perform an ablation study for our regularization term. For this, we report the cosine similarity between the explanatory  and discriminative information through the DVAE, i.e., $\cos\left({\cal I}_\nu (g), {\cal I}_\nu (f) \right)$ where the input of $g$ is $\ReconstructedPhi$ and the input of $f$ is $\phi(x)$. We observe that with our regularization, the shared explanatory and discriminative information is significantly higher.}
\vspace{-0.3cm}
\label{tab:cosine}

\end{table}

\vspace{-0.1cm}
\subsection{Dataset bias}\label{sec:bias_control}
\vspace{-0.05cm}

Bias in facial analysis has gained increasing attention over time~\cite{wang2020towards, xu2020investigating}. Those biases are attributed to spurious correlations in the data. For example, in CelebA, most of the women are smiling while men are not. As a result, classifiers learn to exploit these correlations resulting in biased models. Traditionally, practitioners had to guess the bias of a particular concept and test the classifier accordingly. We reveal those concepts in an unsupervised manner using our method. In the following, we intentionally introduce gender bias into a dataset. We then demonstrate how our method reveals these biases. 

\vspace{-0.05cm}
\subsubsection{Dataset statistics control}
\vspace{-0.05cm}

Following~\citet{wang2020towards}, we investigated the smiling recognition task. We showed that our method depicts feminine features in cases where the classification network predicts `smiling' and muscular features when it predicts `not smiling.'

To further study this bias, we trained two classifiers with different data splits 1) The female majority split, 70\% of smiling women and 30\% of non-smiling men. 2) The male majority split, 70\% of smiling men and 30\% of non-smiling women. Fig.~\ref{fig:dataset_bias} illustrates that our method recognizes the integrated sex concept. In the first split, smiling faces are illustrated as female (i.e., the second column) and non-smiling faces as male (i.e., the fifth column). On the other hand, the opposite occurs with the second split (i.e., the third and sixth columns)

\vspace{-0.1cm}
\subsubsection{Counterfactual verification} 
\vspace{-0.05cm}
Generative adversarial networks have recently enhanced our ability to manipulate images to change their properties and generate realistic images. One example is StyleCLIP~\cite{patashnik2021styleclip}, which enables text-guided manipulations. We employed this technique and created counterfactual samples. We then measured the effect and assessed the classifier predictions. \figref{fig:counterfactual} illustrates that age, color, and gender affect the attractive, big-lips, and smiling classifiers respectively. Thus, we manipulated images according to these concepts. We altered the concepts as follows: 1) Age by the prompts ``old face'' and ``young face'', 2) Skin color by the prompts ``black person'' and ``white person'', and 3) Sex by the prompts ``male face'' and ``female face''. We provide more details in the supplementary.

\begin{figure}[t!]
\includegraphics[width=0.46\textwidth]{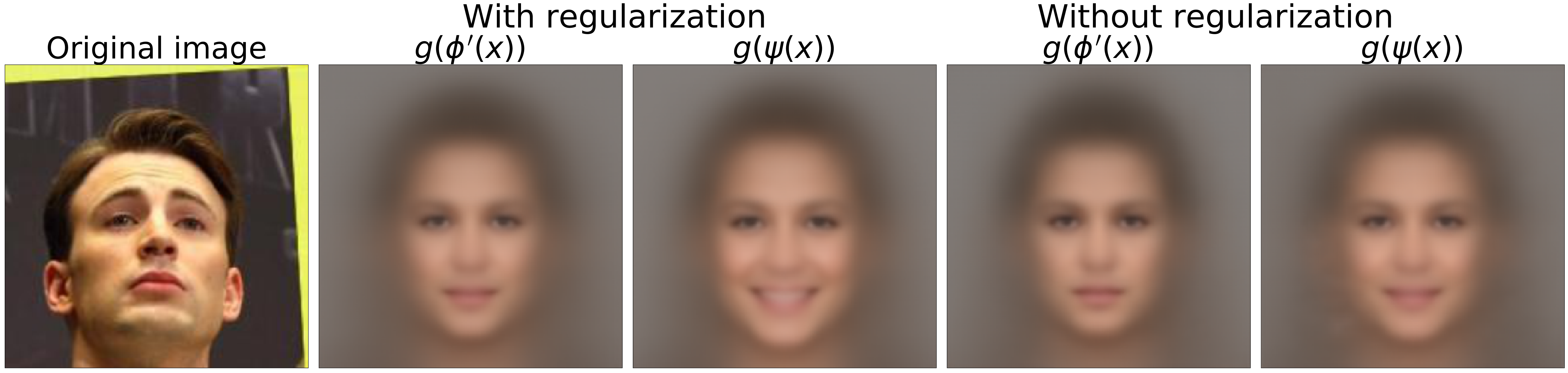}
\centering
\vspace{-0.2cm}
\caption{An illustration of the regularization effect on generated images. A non-smiling person is shown for the smiling task. Our findings indicate that regularization helps to change relevant concepts of prediction through intervention. In the example above, a non-smiling generated image became smiling upon regularization, but it remains non-smiling without regularization.}
\label{fig:regularization}
\vspace{-0.4cm}
\end{figure}

\vspace{-0.1cm}
\subsection{Classifier-generator shared information}
\vspace{-0.1cm}
The main objective of the generator loss is to reconstruct the original image, given a distilled representation of the image. However, the reconstruction loss term alone does not consider to what extent the classifier utilizes each feature of $\phi(x)$. Features in $\phi(x)$ may be relevant for the prediction but not for the reconstruction and vice-versa. In the following, we examined the essential role of our proposed regularization term for generating a reliable explanation. 

\vspace{-0.05cm}
\subsubsection{Quantitative analysis} 
\vspace{-0.05cm}
In this experiment, we measure the cosine similarity between the explanatory and discriminative information of $g$ and $f$ respectively. A cosine similarity metric aims to quantify the similarity between two vectors in the inner product space. This metric is bounded in the range of $[-1, 1]$. Two information vectors that are oriented the same way yield the maximum value of $1$. One can notice that our regularization term is proportional to the cosine similarity. Tab.~\ref{tab:cosine} shows that our regularization term significantly improves the cosine similarity between the explanatory  and discriminative information vectors across tasks.

\subsubsection{Qualitative analysis} 
\figref{fig:regularization} demonstrates the effects of regularization on generated images. Generated images serve to reveal concepts that the classifier employs in making predictions. Thus, interventions should change the concept in a way that also changes prediction. In the smiling task, a non-smiling person is expected, upon intervention, to have a smiling concept image. As we can see on the right, without regularization, the intervention does not result in a smiling figure with teeth. However, on the left, with regularization, an intervention leads to a smiling figure with exposed teeth.

\vspace{-0.1cm}
\section{Conclusion}\label{sec:conclusion}
\vspace{-0.1cm}

Model interpretability plays an essential role in deep neural network debugging. Explanations provide insights about the model's performance rather than just looking at its test-set metrics. It allows a data expert without neural network knowledge to understand why a model made a prediction. By doing so, confidence in such models improves. In this work, we propose an explanatory approach to latent layer explanation. Our proposed method permits the discovery of the underlying concepts a model is using. Furthermore, it allows interventions on those revealed concepts. With our method, we hope to assist researchers in better understanding the operation of their models.

\section{Acknowledgements}

This research was supported by Grant No. 2019783 from the United States-Israel Binational Science Foundation (BSF).

\bibliography{aaai22}

\end{document}